\documentclass[letterpaper]{article} 
\usepackage{aaai2026}  
\usepackage{times}  
\usepackage{helvet}  
\usepackage{courier}  
\usepackage[hyphens]{url}  
\usepackage{graphicx} 
\usepackage{natbib}  
\usepackage{caption} 
\usepackage{algorithm}
\usepackage{algorithmic}

\usepackage{xcolor}
\usepackage{amssymb}
\usepackage{MnSymbol}
\usepackage{makecell}
\usepackage{float}
\usepackage{array}
\usepackage{placeins}
\usepackage{multirow}
\usepackage{tabularx}
\usepackage{todonotes}
\usepackage{booktabs}
\usepackage{ifthen}

\usetikzlibrary{positioning, arrows.meta, shapes.multipart, shapes.geometric, fit, calc, decorations.pathreplacing}

\usepackage{newfloat}
\usepackage{listings}
\DeclareCaptionStyle{ruled}{labelfont=normalfont,labelsep=colon,strut=off} 
\floatstyle{ruled}
\newfloat{listing}{tb}{lst}{}
\floatname{listing}{Listing}
\title{
MArgE: Meshing 
Argumentative Evidence from\\ Multiple Large Language Models for Justifiable Claim Verification}

\author {
    Ming Pok Ng\textsuperscript{\rm 1}\equalcontrib,
    Junqi Jiang\textsuperscript{\rm 1}\equalcontrib\thanks{Correspondence to {junqi.jiang@imperial.ac.uk}. Preprint.},
    Gabriel Freedman\textsuperscript{\rm 1},
    Antonio Rago\textsuperscript{\rm 12},
    Francesca Toni\textsuperscript{\rm 1}
}
\affiliations {
    \textsuperscript{\rm 1}Department of Computing, Imperial College London, UK\\
    \textsuperscript{\rm 2}Department of Informatics, King's College London, UK
}

\usepackage{bibentry}

\begin{document}

\maketitle

\begin{abstract}

Leveraging outputs from multiple large 
language models (LLMs) is emerging as a method for harnessing their 
power across a wide range of tasks while mitigating their capacity for making errors, e.g., 
hallucinations. However, current approaches to combining insights from multiple LLMs often involve unstructured interactions (e.g., free debate), resulting 
in model generations that are not faithfully justifiable. In this work, we introduce MArgE, a novel framework 
to provide formal structure to the 
evidence from each LLM, in the form of a tree of extracted arguments,
for the task of claim verification. We use a variant of Argumentative LLMs (ArgLLMs), i.e. LLMs driven by frameworks and semantics from the field of computational argumentation, to construct structured {argument} trees for given claims. This process creates an inspectable pathway from the initial arguments to the final claim verification decisions, 
providing a faithful justification thereof.
We show experimentally that MArgE
can significantly outperform single LLMs,
including three open-source models (4B to 8B parameters), 
GPT-4o-mini and
existing ArgLLMs, as well as prior methods for unstructured multi-LLM debates. 
We thus demonstrate the advantages of incorporating formal, argumentative reasoning mechanisms 
when combining multiple LLM outputs.

\end{abstract}




\section{Introduction}
\label{sec:introduction}

\begin{figure*}[h!]
\centering
  \includegraphics[width=\textwidth]{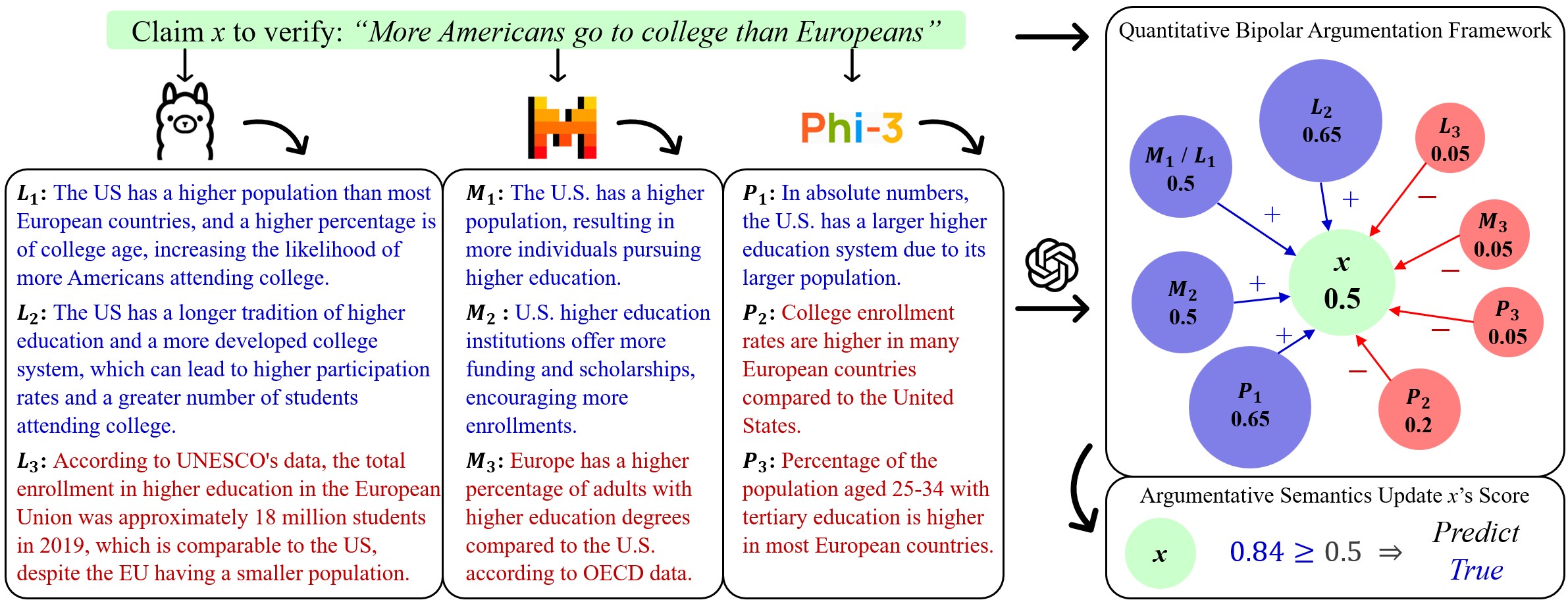}
  \caption{
  An illustrative example of MArgE on the Truthful Claim dataset. First, given the claim, three LLMs (Llama-3.1-8B-Instruct, Mistral-7B-Instruct-v0.3 and Phi-3-Mini-4K-Instruct (3.8B)) generate three arguments for or against the claim, marked respectively in blue and red. Then, an argumentation framework is built, with each argument and the claim rated by GPT-4o-mini with a scalar score. Each new argument points to the claim with a support (+) or attack (-) arrow (note that more complex frameworks with greater depths are possible). By applying argumentative semantics, the final claim strength is calculated as 0.837, resulting in a True prediction, which is correct. In contrast, GPT-4o-mini's CoT output states ``...the U.S. does have a higher percentage of young adults with college degrees... However, this doesn't necessarily mean that more Americans are enrolled... we can conclude that while many Americans do go to college, the claim lacks a definitive basis...'', which expresses complex reasoning and significant uncertainty. This leads GPT-4o-mini to predict False, which is incorrect.
  }
  \label{fig:page1}  
\end{figure*}

Large language models (LLMs) have emerged as powerful tools in deep learning, excelling at tasks such as generating coherent text, answering complex queries and assisting in decision-making across a wide range of domains \cite{DBLP:conf/nips/VaswaniSPUJGKP17,brown2020,DBLP:conf/nips/Ouyang0JAWMZASR22}. 
These models are increasingly being explored for high-risk applications such as claim verification, where they can help identify factual inaccuracies and support automated reasoning \cite{DBLP:conf/emnlp/WangS23a}. 
Methods based on prompting and decoding have been developed to present the internal reasoning traces of a single LLM, thereby improving performance and transparency \cite{wei2022,yao2023,besta2024}. 
To enhance utility and reliability, combining outputs from multiple LLMs has also become a popular line of research \cite{chen2025}. 
Specifically, LLM ensembling, such as debate- and voting-based methods  \cite{du2024,khan2024,estornell2024,DBLP:conf/nips/GuhaCCKR24}, aim to encourage more robust outcomes by aggregating diverse model perspectives. However, despite these improvements in task performance, existing approaches remain limited in justifiability - it is often ambiguous how final decisions are reached. Recent findings show that Chain-of-Thought (CoT) outputs can suffer from hallucinations and may not faithfully reflect the true reasoning process of the models \cite{arcuschin2025chain,barez2025chain}, undermining their reliability as post hoc justifications. Additionally, the unstructured nature of multi-LLM debates makes the rationale behind the collective decisions difficult to trace. Recently, Argumentative LLMs (ArgLLMs) \cite{Freedman_Dejl_Gorur_Yin_Rago_Toni_2025} have been shown to exhibit better faithfulness of the decision process 
than a single LLM. This is achieved via techniques from computational argumentation (see \cite{AImagazine17} for an overview), an area of symbolic AI with principled formalisms for structuring arguments in the form of conflicting evidence.

Inspired by ArgLLMs, in this work, we
propose MArgE, an approach that combines decomposed evidence across multiple LLMs into a unified, interpretable structure. Specifically, instead of asking each model to directly produce CoT outputs to judge the claim, we ask for 
arguments that would either support or attack the claim. A tree (or graph) is then built to connect these arguments and the claim, where the quality of each node is evaluated through another LLM. Finally, computational argumentation semantics are applied to propagate the arguments' dialectical strengths, culminating in the update of 
the claim's strength, which we finally use to predict the verification result. Such an approach ensures that multiple information sources are included, and that the decision process is readily justifiable by inspecting the tree with the arguments' quality scores and strengths. The final prediction is 
thus faithful to the generated arguments and is grounded by argumentative semantics \cite{Freedman_Dejl_Gorur_Yin_Rago_Toni_2025}. Figure \ref{fig:page1} shows an illustrative example of MArgE and its advantages over prompt-based LLM outputs. This example illustrates how MArgE integrates diverse and potentially conflicting perspectives from multiple models, allowing the final verdict to be driven by the structure and strength of evidence rather than any single model's binary judgement.

After covering the preliminaries in Section \ref{sec:preliminaries}, we introduce our method in Section \ref{sec:main}. We present a comprehensive evaluation of the effectiveness of MArgE in Section \ref{sec:evaluation} with three open-source models of varying sizes, a proprietary LLM GPT-4o-mini \cite{openai2024gpt4o}, and three claim verification datasets. We demonstrate that MArgE can significantly outperform strong baselines, including GPT-4o-mini, ensembles of ArgLLMs and multi-LLM debates \cite{khan2024}. We also conduct a detailed ablation study, providing insights into the strengths of MArgE. We then review related literature in Section \ref{sec:related} and finally conclude in Section \ref{sec:conclusions}.



\section{Preliminaries}
\label{sec:preliminaries}



\paragraph{Claim Verification with Multiple LLMs} We characterise the scenario targeted by this work as the following. Given a claim $x\in \mathcal{X}$ where $\mathcal{X}$ denotes the space of all possible natural language sequences, a finite set of $K>1$ LLMs $\mathcal{G}=\{G_1, \ldots, G_K\}$ is instructed to collectively determine whether the claim $x$ is \textit{true} or \textit{false}. Each $G_i \in \mathcal{G}$ is such that $G_i: \mathcal{X}\rightarrow\mathcal{X}$. For simplicity, we use a function $f_{\mathcal{G}}:\mathcal{X}\rightarrow \{0,1\}$ to represent the claim verification process by $\mathcal{G}$, where label $1$ ($0$) indicates that the claim is true (false, respectively). Practically, $f$ could include the specific prompt construction over $x$ and any intermediate outputs from each LLM, sampling for LLM responses, and parsing for final predictions.

\paragraph{Quantitative Bipolar Argumentation 
}
A \emph{quantitative bipolar argumentation framework (QBAF)}~\cite{baroni2019} is a quadruple $\langle A,R^{-},R^{+},\tau\rangle$ with
\begin{itemize}
  \item A finite set of \emph{arguments} $A$.
  \item Disjoint binary relations of \emph{attack} $R^{-}\subseteq A\times A$ and \emph{support} $R^{+}\subseteq A\times A$.
  \item A \emph{base-score} (intrinsic strength) function $\tau:A\to[0,1]$.
\end{itemize}
%
%
Then, 
\emph{gradual semantics} 
recursively compute an argument's \emph{dialectical strength} by combining 
its base score with the aggregated 
strengths of its attackers and supporters.
Given a gradual semantics
, such as \emph{DF-QuAD} \cite{rago2016}, denoted $\sigma_Q$, each argument $\alpha\in A$ obtains a 
strength
$\sigma_Q(\alpha)\in[0,1]$\footnote{Let $v_0=\tau(\alpha)$, 
$S^{-}(\alpha)$ ($S^{+}(\alpha)$) be an arbitrary permutation of the strengths of the attackers (supporters, respectively) of $\alpha$,
$v_a=F(S^{-}(\alpha))$ and
$v_s=F(S^{+}(\alpha))$,
then, $\sigma_Q(\alpha)=C(v_0,v_a,v_s)$, where, as defined in~\citet{rago2016}: 
$F(())\!=\!0$ and, for $v_1,\ldots,v_n \!\in \![0,1]$ ($n \geq 1$),
if $n=1$ then $F((v_1))=v_1$,
if $n=2$ then $F((v_1,v_2))=
v_1 + v_2 - v_1\cdot v_2$,
and if $n>2$ then
$F((v_1,\ldots,v_n)) = F(F((v_1,\ldots, v_{n-1})),v_n)$; and
if $v_a\geq v_s$ then $C(v_0,v_a,v_s)=v_0-v_0\cdot\mid v_s - v_a\mid$ and
if $v_a< v_s$, then
$C(v_0,v_a,v_s)=v_0+(1-v_0)\cdot\mid v_s - v_a\mid$.}.

\paragraph{Argumentative Large Language Models
} ArgLLMs \cite{Freedman_Dejl_Gorur_Yin_Rago_Toni_2025}
augment an LLM with a four-stage pipeline that
constructs and evaluates a QBAF for explainable single-LLM claim verification. This can be viewed as a special case of our problem setting when 
setting $K=1$, i.e., $\mathcal{G}=\{G\}$. Consider an input $x$, an ArgLLM operates as:

\begin{enumerate}
  \item \textbf{Argument generation} 
  $\Gamma_{\mathcal{G}}(x)\!\rightarrow\!B$ uses the LLM to produce (guided by hyperparameters such as tree depth) a bipolar argumentation \emph{tree} $B$ whose root is $x$. Every other node is an argument generated by $G$. Every node, apart from the leaf nodes, has exactly one newly generated attacker and one supporter argument pointing to it.
  \item \textbf{Intrinsic argument strength attribution} $\mathcal{E}_{E}(B)\!\rightarrow\!Q$ 
        assigns a base score to every node via some evaluator model $E$ ($E=G$, i.e., the LLM itself provides the base score on its arguments, in original ArgLLM instantiation), yielding a QBAF $Q$.
  \item \textbf{Argumentative strength calculation} $\Sigma_{\sigma}(Q)\!\rightarrow\!\sigma_Q(x)$  
        applies a gradual semantics $\sigma$ to compute the dialectical strength of every argument, in particular the root claim.
  \item \textbf{Claim verification prediction} $g(\sigma_Q(x)) \rightarrow \{0, 1\}$ predicts the final result - the claim is true when $\sigma_Q(x) \!\geq\! 0.5$ or false otherwise. 
\end{enumerate}

Overall, ArgLLM prediction can be expressed as the composition of the above operations, denoted as $f_{\textit{ArgLLM}}(x)=g\,\circ\,\Sigma_{\sigma}\,\circ\,\mathcal{E}_{E}\,\circ\,\Gamma_{\mathcal{G}}$.
    


\section{MArgE: Meshing Argumentative Evidence from Multiple LLMs} 
\label{sec:main}

\begin{figure*}[ht]
    \centering
    \resizebox{0.9\textwidth}{!}{\input{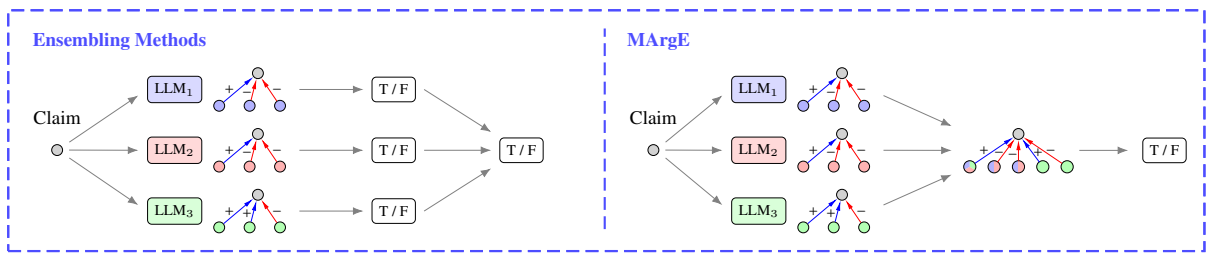}}
    \caption{Comparison of simple ensembling of predictions from separate argumentative trees (left) against MArgE (right), where a claim is independently processed by multiple LLMs to generate structured pro/con argument trees. These trees are then merged into a unified structure, scored for intrinsic strength, and evaluated using gradual semantics to produce a final prediction. 
    It can thus be seen how MArgE meshes argumentative evidence in targeting increased justifiability.
    }
    \label{fig:pipeline}
    \end{figure*}

We now introduce 
MArgE, 
our novel framework that leverages computational argumentation to 
mesh the generated evidence from multiple LLMs. Figure \ref{fig:pipeline} presents the overall pipeline, which proceeds in four steps as follows.

\paragraph{Step 1: Supporting and attacking argument tree generation} Given a base claim $x$, each LLM $G_i \in \mathcal{G}$ is first independently prompted to generate a tree $B_i$ of \textit{pro} (supporting) and \textit{con} (attacking) arguments for the claim, forming the evidence from each $G_i$. The tree structure and depth are customisable. Specifically, starting with the root argument, $x$, the first layer of $B_i$ consists of up to three arguments, each supporting or attacking $x$. The subsequent layers then evaluate every node argument of its previous layer, for which another (at most) three pro and/or con arguments are generated again. This process terminates at the specified layer. Each resulting tree structurally reflects the reasoning process of its corresponding model towards $x$.

For each node in the tree, the LLM is instructed to choose the number of arguments generated (up to three in our instantiations) and their relation with the parent node, allowing models to articulate their stances more comprehensively and capture a wider range of reasoning. This enables the integration of a broader spectrum of argumentative perspectives. 
The resulting trees are structured as Bipolar Argumentation Frameworks (BAFs)~\cite{cayrol2005}, 
i.e., QBAFs without base scores. We denote this step as $\Gamma_{\mathcal{G}}(x)=\mathcal{B}$, where 
$\mathcal{B}=\{B_1, \ldots, B_K\}$ is a set of BAFs.

\paragraph{Step 2: Meshing multiple argument trees} To perform argumentative reasoning across evidence from multiple LLMs via gradual semantics, we mesh the argument trees into a single one. 
One basic approach is to concatenate all the arguments of every tree at each layer, apart from the root node, which is fixed to $x$. We refer to this as \textit{simple union}. 

Although it preserves argument diversity, simple union may introduce redundancy. To take into account the potentially similar arguments generated by different models, we further allow meshing tree nodes based on their semantic similarity. Specifically, this step involves a lightweight, encoder-only, sentence transformer model \cite{sentence-bert}, denoted as $\mathcal{S}: \mathcal{X}\rightarrow \mathbb{R}^d$, where $d$ is the dimension of its embedding space. The output of $\mathcal{S}$ encodes the semantic meaning of the input text. Then, the semantic similarity between a pair of texts $\alpha_1, \alpha_2 \in \mathcal{X}$ can be calculated as $cosine\_similarity(\mathcal{S}(\alpha_1), \mathcal{S}(\alpha_2))$. We then calculate the similarity between each pair of arguments from different trees, and merge the pair if the similarity is above some threshold. We refer to this option as \textit{semantic merging}. This step is denoted as $\mathcal{M}(\mathcal{B})=B$.



\paragraph{Step 3: Argument quality scoring}

After meshing, we transform the resulting argument tree into a QBAF by assigning a base score (between 0 and 1) to each argument, representing the strength of an argument independent of its attackers and supporters. The scores are determined based on prompting an LLM, denoted $E:\mathcal{X}\rightarrow[0,1]$. Differently to the original ArgLLMs \cite{Freedman_Dejl_Gorur_Yin_Rago_Toni_2025}, which use the argument-generating model for providing their \textit{confidence} as the 
base scores, we use an external LLM for this purpose in a reward model style \cite{DBLP:conf/nips/StiennonO0ZLVRA20,bai2022training,DBLP:conf/naacl/LambertPMMLCDKZCSH25} presenting both the argument and its parent node to the scorer, asking for a \textit{quality} score based on relevant aspects. This is to incorporate more diverse information sources and thus mitigate systematic bias in the framework. We include two scoring approaches for the merged tree $B$, referred to as $\mathcal{E}_{E}(B)=Q$:

\begin{itemize}
    \item \textbf{Estimated Arguments:} The root claim $x$ is assigned a fixed score of 0.5, and the generated arguments are scored by $E$. This is done by asking the scorer model to consider the \emph{relevance}, \emph{factuality} and \emph{ambiguity} of each argument with respect to its parent node, then choose from seven discrete labels ranging from ``certain'' to ``uncertain'', which are then converted into the interval $[0, 1]$ through a fixed mapping (See Appendix A for detail). 
    This setup allows us to aggregate perspectives without biasing the outcome with a predefined root confidence.
    
    \item \textbf{Estimated All:} 
    Extending the previous approach, this option additionally evaluates the claim $x$. The scorer is asked to examine the \textit{truthfulness} of the claim and assign it with one of five discrete labels ranging from ``true'' to ``false'', which are then converted into the interval $[0, 1]$ through a fixed mapping. This configuration 
    leverages the evaluator model's judgements on both the claim itself and its supporting/attacking arguments, providing richer signals for argumentative reasoning. 
\end{itemize}


\paragraph{Step 4: Updating argument strength with gradual semantics} The base scores in the QBAF $Q$ reflect the quality of each argument. We finally apply gradual semantics $\sigma$ (such as DF-QuAD~\cite{rago2016}) to propagate the influence of every argument through the graph. During the process, each node's strength is updated iteratively based on its base score and its 
attackers and supporters. We can thus obtain an updated score for the root node, $\sigma_Q(x)$, which is then thresholded at 0.5 for the final true/false prediction.

We further explored the option to mine additional relations across each model's argument tree with a fine-tuned LLM for tree merging, and tested different argumentation semantics for updating argument strength. These are included in Appendix C. Combining all steps, MArgE can be expressed as the composition of the above operations, denoted as 
$f_{\textit{MArgE}}(x)=g\,\circ\,\Sigma_{\sigma}\,\circ\,\mathcal{E}_{E}\,\circ\,\mathcal{M}\,\circ\,\Gamma_{\mathcal{G}}$. In contrast to the ArgLLM methodology, we use different methods for $\Gamma_{\mathcal{G}}$ and $\mathcal{E}_{E}$, with a new meshing step $\mathcal{M}$.

\paragraph{Discussion on Justifiability} Justifiability reflects the degree to which a decision process can be understood, rationalised, and grounded in transparent reasoning, which our proposed framework aims to address \cite{DBLP:journals/ai/Miller19}. Traditionally, in prompt-based generation for a single LLM, the model output might not have easily decomposable reasoning traces, and can contain conflicting arguments within the chunk of text, making it difficult for an external agent (either a human or an LLM) to understand, e.g., the CoT outputs from GPT-4o-mini in Figure \ref{fig:page1} example. This approach exhibits a high risk of hallucination, and the outputs might not be faithful to the model's internal reasoning \cite{arcuschin2025chain,barez2025chain}. Any unstructured LLM output aggregation framework could also inherit this limitation. 

MArgE, on the other hand, does not rely on CoT-style outputs to reach a final decision. Given the claim, our approach of obtaining evidence from multiple models, composed as trees of short supporting and attacking arguments, discretises the reasoning and thus provides a diverse and transparent rationale for the claim. Unlike CoT outputs, our merged and scored argument tree (in the form of QBAFs) is intuitive to inspect. The external scoring step for each argument also marks down the obviously hallucinated arguments such that they will not have a strong impact towards the updated scores of the claim (e.g., argument L3 and M3 are assigned with 0.05 scores in Figure \ref{fig:page1} example). The discrete labels for rating the arguments also provide human-interpretable signals to the QBAF. Further, the existing multi-LLM debate methods are 
also vulnerable to the involvement of adversarial agents misleading the debate process \cite{Amayuelas2024MultiAgentCA}, whose arguments might be readily marked down in our framework. We leave the relevant investigation for future work. Finally, our use of formal 
argumentative semantics, e.g., DF-QuAD, ensures that the final verdict is not only faithful to the transparent reasoning traces presented in the QBAF, but also logically consistent with formal groundings in computational argumentation theories \cite{amgoud2018,baroni2019}. 



\section{Evaluation}
\label{sec:evaluation}

We now assess MArgE's accuracy in the claim verification task with an empirical evaluation.

\subsection{Experiment Setup} \label{ssec:experimental-setup}



\paragraph{Datasets} We evaluate MArgE on three binary claim verification datasets: TruthfulClaim, StrategyClaim, and MedClaim. These datasets were originally curated by \citet{Freedman_Dejl_Gorur_Yin_Rago_Toni_2025} and were derived from the TruthfulQA~\cite{truthfulqa}, StrategyQA~\cite{strategyqa}, and MedQA~\cite{medqa} datasets. These datasets provide broad coverage across different reasoning competencies that are central to trustworthy LLM behaviour, respectively targeting resistent to common misconceptions, multi-hop inference, and domain-specific knowledge. Each dataset is constructed by transforming QA pairs into standalone factual claims with binary truth labels. This transformation was performed using LLM-assisted rewriting followed by human verification to ensure grammaticality, factuality and clarity. We use a held-out validation split of 200 claims from each dataset for exploratory experiments for MArgE, then use different test sets for evaluation against baselines, each containing 250 true and 250 false claims.

\paragraph{Models} MArgE leverages an ensemble of LLMs, each independently generating structured pro/con argument trees. 
We use Llama-3.1-8B-Instruct~\cite{llama}, Mistral-7B-Instruct-v0.3~\cite{mistral} and Phi-3-Mini-4K-Instruct~\cite{phi} (3.8B). These models cover a diverse and competitive set of architectures and training paradigms, spanning different parameter scales and instruction-following capabilities. 
All models are publicly available on Hugging Face~\cite{huggingface}. We additionally use a stronger model, OpenAI's GPT-4o mini (GPT-4o-mini-2024-07-18)~\cite{openai2024gpt4o}, to assign a quality score for the arguments and claims. 

To enable efficient inference across multiple models simultaneously without noticeable performance degradation, we use 4-bit quantised versions of each LLM via QLoRA~\cite{qlora}. 
All LLMs were queried with 0.7 temperature and top-p of 0.95 for balanced randomness. See Appendix A for details on prompting LLMs. All experiments were run on a GPU cluster, comprising a mix of NVIDIA Tesla A40 (48 GB VRAM) and A30 (24 GB VRAM) nodes, each paired with AMD EPYC processors.





\paragraph{Baselines} We compare the 
performance of MArgE against the following baselines:
\begin{itemize}
    \item \textbf{Single LLM} (direct questioning and CoT prompting): We individually assess the argument trees produced by the three open-source LLMs (Llama, Mistral, Phi).
    \item \textbf{Ensemble of LLMs}: An ensemble baseline that takes the majority vote prediction of the above three models. This baseline allows a direct comparison against our proposed MArgE, which also aggregates outputs from the same models.
    \item \textbf{Ensemble of ArgLLMs}: This new adapted baseline takes the majority vote prediction from the ArgLLM instantiations with each open-source LLM. This is illustrated in Figure \ref{fig:pipeline} (left). During the QBAF construction, unlike in the original ArgLLM work, we employ the same argument tree generation (Step 1) and argument quality scoring (Step 3) approaches as in our proposed MArgE, allowing for a fair comparison. We additionally compare with single-model ArgLLM in Appendix B.
    \item \textbf{Debate} \cite{khan2024}: This method 
    involves two debating LLMs, one defending and the other countering the ground truth, with another external judge LLM reading the debate script and concluding a final verdict. We experiment with 3 rounds of debates and GPT-4o mini as the judge. We instantiate Debate between all pairwise combinations of our three open-source LLMs. For each model pair, we run four configurations where each model alternates as the starter, defender and counter. Each configuration is denoted by the first letters of the models taking each role, e.g., \textit{Debate,MLM} (Mistral starts, Llama defends, Mistral counters).
    \item \textbf{GPT-4o mini} (direct questioning and CoT prompting): This model is not involved in the argument generation, but is used to rate each argument with a score. Including it here allows us to establish a performance baseline for the underlying model used in scoring.
\end{itemize}

Note that we experiment with both direct questioning and CoT prompting \cite{wei2022,wang2023} for the single LLM baselines and GPT-4o mini. 
We report the higher accuracy results over two prompting methods for every baseline. See Appendix A and Appendix B for detailed prompts and results.

We instantiate MArgE with depth 1 (D1) and depth 2 (D2) for Step 1, simple union and semantic merging for Step 2, estimated arguments (Est.Arg) and estimated all (Est.All) for Step 3. Step 1 and Step 3 configurations also apply to the ensemble of ArgLLMs baseline. We report the mean result of three separate runs for these two methods.

\subsection{Accuracy Results}
\label{ssec:accuracy_results}

\begin{table}[t]
\centering
\begin{tabularx}{\columnwidth}{c|ccc}
\hline
\toprule
\textbf{Method} & \textbf{Truth} & \textbf{Strategy} & \textbf{Med} \\ \midrule
 {Llama-3.1-8B-Inst.} & 0.710 & 0.652 & 0.564 \\
 {Mistral-7B-Inst.} & 0.702 & 0.604 & 0.554 \\
 {Phi-3-Mini-Inst.(4B)} & 0.656 & 0.612 & 0.602 \\
 \midrule
 {Ensemble of LLMs} & 0.724 & 0.642 & 0.570 \\

 {Ens.ArgLLM-D1-Est.Arg} & 0.683 & 0.661 & 0.568 \\
 {Ens.ArgLLM-D2-Est.Arg} & 0.679 & 0.669 & 0.564 \\
 \textbf{MArgE-D1-Est.Arg} & {0.715} & 0.695 & 0.580 \\
 \textbf{MArgE-D2-Est.Arg} & 0.689 & 0.685 & 0.574 \\

 \midrule
 {GPT-4o-mini} & 0.802 & 0.794 & \textbf{0.760} \\
 \midrule
 {Ens.ArgLLM-D1-Est.All} & 0.823 & 0.782 & 0.682 \\
 {Ens.ArgLLM-D2-Est.All} & 0.811 & 0.785 & 0.669 \\
 \textbf{MArgE-D1-Est.All} & \textbf{0.837} & 0.797 & 0.715 \\
 \textbf{MArgE-D2-Est.All} & \textbf{0.837} & \textbf{0.804} & 0.733 \\
\bottomrule
\end{tabularx}

\caption{Accuracy comparison against the three single LLMs, ensemble of LLMs, ensemble of ArgLLMs, and GPT-4o mini baselines. Rows above the GPT-4o mini baseline do not use this model to rate the claim, while rows below it do. MArgE is instantiated with three open-source models and GPT-4o mini (as the base scorer). The highest accuracy for each dataset is highlighted in bold. Our method reports the best of union and semantic merging.}


\label{tab:result-main}
\end{table}

We report the main accuracy results in Table \ref{tab:result-main}, comparing MArgE instantiated with three open-source LLMs against the baselines. We separately report accuracy and computational cost comparison between MArgE and Debate, both with two LLMs, in Tables \ref{tab:result-debate} and \ref{tab:result-tokens}. Overall, MArgE often achieves better performance than the baselines.

\paragraph{General Comparison} First, we focus on the first two blocks of Table \ref{tab:result-main}, where no methods use the stronger GPT model to rate the root claim. MArgE, when using depth of~1, outperforms all baselines above on at least two datasets, demonstrating promising performance gains. It also has the highest performance on the Strategy Claim dataset. 
The ensembles of LLMs and ArgLLMs do not improve over the most accurate single model on two of three datasets. Indeed, given the disparity in the three model's performances, it is possible that the ensembled prediction is governed by the less accurate models. MArgE improves on this aspect, highlighting the usefulness of meshing the diverse evidence from multiple models.

Significant accuracy improvements over the above methods (up to around $20\%$) can be observed when allowing the stronger GPT-4o mini model to be more involved (the last two blocks of Table 1). Notably, MArgE, without directly prompting the scorer for an answer over the claim, outperforms the GPT scorer's CoT accuracy on two of three datasets by $3.5\%$ and $1.0\%$. On MedClaim, MArgE is less accurate, possibly due to the fact that substantial domain-specific knowledge is extracted from the open-source models as part of the argument tree, which they are less specialised in. MArgE also consistently outperforms the ensemble of ArgLLMs, highlighting the effectiveness of meshing.

Interestingly, we found that with the Est.Arg option, MArgE performs better with an argument depth of 1. Accuracy is higher with depth 2 when additionally estimating the claim. We will discuss this in detail in Section \ref{ssec:ablation}.

\begin{table}[h]
\centering
\resizebox{\columnwidth}{!}{
\begin{tabular}{ccccc}
\toprule
\textbf{Models} & \textbf{Method} & \textbf{Truth} & \textbf{Strategy} & \textbf{Med} \\
\midrule
\multirow{5}{*}{\shortstack[c]{Llama\\+\\Mistral}} 
 & Debate,LLM & 0.760 & 0.664 & \underline{0.654} \\
 & Debate,LML & 0.768 & 0.716 & 0.564 \\
 & Debate,MLM & \underline{0.808} & \underline{0.738} & 0.638 \\
 & Debate,MML & 0.676 & 0.668 & 0.630 \\
 & \textbf{MArgE-D1} & \textbf{0.814} & 0.798 & 0.668 \\
 & \textbf{MArgE-D2} & {0.806} & \textbf{0.800}& \textbf{0.688} \\
\midrule
\multirow{5}{*}{\shortstack[c]{Llama\\+\\Phi}} 
& Debate,LLP & 0.822 & 0.762 & \underline{0.664} \\
& Debate,LPL & 0.684 & 0.640 & 0.494 \\
& Debate,PLP & \textbf{\underline{0.842}} & \underline{0.796} & 0.630 \\
& Debate,PPL & 0.594 & 0.568 & 0.552 \\
&\textbf{MArgE-D1} & {0.818} & 0.780 & 0.718 \\
& \textbf{MArgE-D2} & {0.816} & \textbf{0.798} & \textbf{0.722} \\
\midrule
\multirow{5}{*}{\shortstack[c]{Mistral\\+\\Phi}} 
& Debate,MMP & 0.810 & 0.730 & 0.630 \\
& Debate,MPM & 0.672 & 0.634 & 0.612 \\
& Debate,PMP & \textbf{\underline{0.838}} & \underline{0.772} & 0.582 \\
& Debate,PPM & 0.606 & 0.574 & \underline{0.638} \\
& \textbf{MArgE-D1} & 0.826 & 0.780 & 0.688 \\
& \textbf{MArgE-D2} & {0.824} & \textbf{0.792} & \textbf{0.722} \\
\bottomrule
\end{tabular}
}
\caption{Accuracy comparison against the Debate baseline. 
Both MArgE (Est.All) and Debate are instantiated with two open-source LLMs at a time. The highest accuracy is highlighted in bold and the best configuration among Debate instantiations is underlined, for each model combination and dataset. Our method reports the best of union and semantic merging. 
}
\label{tab:result-debate}
\end{table}

\begin{table}[h]
\centering
\begin{tabular}{ccccc}
\toprule
\textbf{Method} & \!\!\textbf{Model}\!\! & \textbf{\shortstack[c]{Llama\\+Mistral}} & \textbf{\shortstack[c]{Llama\\+Phi}} & \textbf{\shortstack[c]{Mistral\\+Phi}} \\
\midrule
\multirow{2}{*}{Debate} 
& GPT & 524/250 & 490/221 & 456/241 \\
& Open & 1727/284 & 1709/294 & 1810/277 \\
\midrule
\!\!\multirow{2}{*}{MArgE-D1}\!\! 
& GPT\textsuperscript{$\dag$} & 4018/105 & 3829/100 & 3644/95 \\
& Open & 514/190 & 634/179 & 619/176 \\
\midrule
\!\!\multirow{2}{*}{MArgE-D2}\!\!
& GPT\textsuperscript{$\dag$} & \!\!13059/341\!\! & \!\!12444/325\!\! & \!\!11844/309\!\! \\
& Open & 1757/743 & 2222/708 & 2078/674 \\
\bottomrule
\end{tabular}
\caption{Computational cost comparison against the Debate baseline for each model combination, averaged across three datasets. GPT and Open, respectively, report the tokens used for the GPT-4o mini model, and the sum of the two open-source models. Two numbers in each entry stand for the amount of input/output tokens. \textsuperscript{$\dag$} Our implementation prompts GPT for base scores separately for each argument, which could be simplified and result in fewer tokens.}
\label{tab:result-tokens}
\end{table}

\paragraph{Accuracy and Cost Comparison with Debate} In this set of comparisons, both MArgE and Debate use two open-source LLMs for generating the rationales and the GPT model for higher-level evaluation. A difference is that the final result in MArgE is determined by formal argumentative reasoning, whereas in Debate, it is decided by GPT as a judge. We observe from Table \ref{tab:result-debate} that MArgE-D2 outperforms all configurations of Debate in Strategy and Med datasets, and MArgE-D1 remain competitive with the highest accuracy in the Truth dataset with Llama and Mistral. MArgE-D1 also outperforms all Debate setups in 6 out of 9 settings. 
Compared with the best Debate setup, 
MArgE often show improved accuracies by large margins of up to $8.4\%$, and is at most $2.6\%$ 
less accurate on the two cases where Debate performs best.

Also, Debate is sensitive to the LLMs' ordering, and the choice of which LLM defends and counters the claim. As we will see in Table \ref{tab:ablation_results} in Section \ref{ssec:ablation}, results from MArgE are more consistent.
Compared with the GPT-4o mini results (Table \ref{tab:result-main}), it is also obvious that the Debate method can easily cause the GPT performance to degrade, whereas MArgE 
consistently shows better accuracy in two of three datasets. 

We report in Table \ref{tab:result-tokens} the number of tokens (for each claim) needed to obtain the results in Table \ref{tab:result-debate}, averaged across datasets. MArgE-D1 uses about half of the tokens of Debate for open-source LLM and GPT-4o mini outputs, about one third for the open-source LLM input tokens, but about eight times the GPT input tokens. Overall, MArgE requires more input tokens but fewer output tokens. We note that in our implementations, we use separate prompts for GPT to score each argument, causing large duplications in input tokens. Should this be implemented in parallel, i.e., score multiple arguments with one prompt, we could observe a 70\% reduction in the number of input tokens for GPT, making our D1 configurations less costly than Debate. However, mild performance drops could be envisaged. While achieving a slightly better performance (about $1\%$, see Table \ref{tab:summarised_ablation}), D2 requires substantially more tokens than D1.

\subsection{Ablation Study}
\label{ssec:ablation}

\begin{table*}[ht]
    \centering


    \begin{tabular}{cccccccc}
        \toprule
        \textbf{Depth} & \textbf{Meshing Strategy} & \textbf{Base Score} & \textbf{Truthful} & \textbf{Strategy} & \textbf{Med} \\
        \midrule
        1 & Union    & Est.Arg   & 0.715 $\pm$ 0.019 & 0.695 $\pm$ 0.011& 0.580 $\pm$ 0.014\\
        1 & Union    & Est.All    & \textbf{0.837} $\pm$ 0.005 & \underline{0.797} $\pm$ 0.008& \underline{0.715} $\pm$ 0.016\\
        1 & Semantic & Est.Arg  & 0.691 $\pm$ 0.006 & 0.682 $\pm$ 0.010 & 0.580 $\pm$ 0.012\\
        1 & Semantic & Est.All   & \underline{0.826} $\pm$ 0.012 & \underline{0.788} $\pm$ 0.009& \underline{0.715} $\pm$ 0.017 \\
        2 & Union    & Est.Arg  & 0.689 $\pm$ 0.008 & 0.685 $\pm$ 0.013&  0.574 $\pm$ 0.009\\
        2 & Union    & Est.All   & \textbf{0.837} $\pm$ 0.001 & \textbf{0.804} $\pm$ 0.011& \textbf{0.733} $\pm$ 0.008\\
        2 & Semantic & Est.Arg & 0.674 $\pm$ 0.014 & 0.677 $\pm$ 0.015  & 0.571 $\pm$ 0.011 \\
        2 & Semantic & Est.All  & \underline{0.828} $\pm$ 0.008 & \underline{0.802} $\pm$ 0.019 & \underline{0.728} $\pm$ 0.009\\
        \bottomrule
    \end{tabular}

    \caption{
        Accuracy across configurations (bold indicates best performance, underlines mark values within 0.03 of the best).
    }
    \label{tab:ablation_results}
\end{table*}

\begin{table*}[h]
    \centering
    \begin{tabular}{cccc}
    \toprule
        \textbf{{Mesh: Union$\rightarrow$Semantic}} & \textbf{{Base Score: Est.Arg$\rightarrow$Est.All}} & \textbf{{Depth when Est.Arg: 1$\rightarrow$2}} & \textbf{{Depth when Est.All: 1$\rightarrow$2}} \\ 
        \midrule
        $-$0.82\%$\pm$0.67\% & +13.31\%$\pm$1.83\% & $-$1.22\%$\pm$0.73\% & +0.90\%$\pm$0.65\% \\
    \bottomrule
    \end{tabular}
    \caption{Average effects on accuracy of each configuration summarised.}
    \label{tab:summarised_ablation}
\end{table*}

Next, to provide more insights into why MArgE demonstrates competitive performance, we explore how each configuration choice affects the accuracy. 
We run MArgE three times on the test sets and also report the variation in accuracy. The results are presented in Table \ref{tab:ablation_results}. We further summarise the average effects of each change in Table \ref{tab:summarised_ablation}. Appendix C also reports ablation results with additional LLM-based relation mining \cite{cabessa2025} across each model's argument tree for tree merging, and a different argumentation semantics \cite{potyka2018} for updating argument strengths.

First, we observe that MArgE, regardless of the configuration, performs consistently against the sampling variability in all LLMs involved (2 open-source LLMs and GPT-4o mini). The largest standard deviation across three runs is 0.019, observed at depth 1, simple union for meshing, estimating new for Truthful Claim dataset and depth 2, semantic merging, estimating all for Strategy Claim dataset. The remaining configurations all have about $1\%$ accuracy variation. This level of stochastic stability suggests that MArgE is suitable for real-world deployment scenarios where reproducibility is essential. 

As summarised in Table \ref{tab:summarised_ablation}, the most influential choice is when we use the scorer to estimate the quality of the claim in addition to the newly generated arguments. By doing this, a consistent performance gain of about $13.31\%$ can be observed. The two 
meshing strategies do not show obvious performance differences. Further, deeper trees may propagate useful signals when the root claim is properly scored, and can also introduce instability when the root is fixed.

\section{Related Work}
\label{sec:related}




Ensemble methods offer an alternative to single-model inference by leveraging multiple LLMs to compensate for individual model weaknesses and enhance overall output quality~\cite{chen2025}. These methods are broadly categorised into ensemble-before-inference, ensemble-during-inference, and ensemble-after-inference. 
Ensemble-before-inference methods assign an input to one certain LLM via routing \cite{srivatsa2024harnessing,DBLP:conf/iclr/DingM0SMRLA24}. Ensemble-during-inference techniques, e.g., DEEPEN~\cite{Huang2024EnsembleLF}, EVA~\cite{Xu2024BridgingTG}, perform token-level collaboration between LLMs by creating shared model vocabularies in order to aggregate next-token probabilities during decoding.
In contrast, MArgE operates at the level of argument frameworks built around multiple model generations, and thus falls under the classification of ensemble-after-inference. MArgE aligns with the `non-cascade' sub-type~\cite{chen2025}, where ensemble occurs after all generations are complete. Methods in this category are usually voting-based \cite{DBLP:conf/emnlp/SiSZZB23,DBLP:conf/nips/GuhaCCKR24}, and could additionally involve answer regeneration \cite{jiang-etal-2023-llm}.


LLM debate is another important subset of methods for enhancing consistency and reliability in multiple LLM outputs~\cite{du2024, khan2024, estornell2024}. 
Debates offer a powerful mechanism for refining reasoning and eliciting truthful answers from LLMs 
by framing reasoning as iterative critique and exchange, allowing agents to collaboratively refine their reasoning processes ~\cite{du2024}.
Furthermore, incorporating non-expert judges to evaluate debate transcripts or automating this oversight with LLM judges enhances the scalability and reliability of this approach~\cite{khan2024}.

However, debates face several challenges. They remain computationally intensive and are often constrained by limitations in processing extensive debate inputs~\cite{du2024}. They are also susceptible to echo chambers, where shared misconceptions among agents can amplify errors ~\cite{estornell2024}. The involvement of an adversarial agent could also easily mislead the debates \cite{Amayuelas2024MultiAgentCA}. 
Furthermore, the lack of a symbolic reasoning component means that the use of the debate exchange as justifications for the final decision suffers from the same lack of reliable faithfulness as traditional CoT methods \cite{lanham2023measuring,arcuschin2025chain,barez2025chain}. This would be especially the case when an external LLM judge evaluates and concludes the whole debate process \cite{khan2024}. 


\section{Conclusions and Future Work}
\label{sec:conclusions}

In this paper, we present MArgE, a framework that meshes evidence from multiple LLMs into a single, principled argumentative structure. By grounding disagreement of LLM arguments in quantitative bipolar argumentation and propagating strength with gradual semantics, the framework inherently addresses the justifiability of the collective prediction with more transparent reasoning traces: every model’s pro/con stance is preserved, and final verdicts follow an explicit semantics-guided propagation instead of opaque numerical voting or black-box LLM summarisation. We show through comparative experiments that MArgE improves accuracy over strong baselines.



While MArgE targets claim verification, extensions may include a broader range of settings (e.g., multi-class classification, supervised short-form question-answering, and open-ended tasks), tighter integration of domain knowledge with retrieval augmented generation \cite{DBLP:conf/nips/LewisPPPKGKLYR020}, more flexible inference through adaptive depth, and human evaluation of the explanation quality. Incorporating inter-LLM debate with MArgE would also be an exciting path. 


\newpage
\bibliography{bib}

\newpage
\clearpage
\appendix

\section{Model Prompting}
\label{app:model_prompting}

Table \ref{tab:prompts} shows the prompts we use for the single LLM baselines in our experiments. Tables \ref{tab:argument-generation-prompts-llama} to \ref{tab:argument-generation-prompts-phi} list the prompts we use for Step 1 argument tree generation in MArgE. Table \ref{tab:aric-relation-prompt} is the prompt for the additional argument mining step for Step 2 meshing in MArgE, which we discuss in Appendix \ref{app:additional_ablation}. Table \ref{tab:estimated-score-prompts} shows the prompts for Step 3 argument quality scoring in MArgE. Table \ref{tab:label-score-mapping} shows the mapping from the discrete scoring labels to numerical base scores.


\begin{table}[h]
\centering
{\scriptsize
\renewcommand{\arraystretch}{1.4}
\begin{tabular}{p{0.22\linewidth} p{0.70\linewidth}}
\toprule
\textbf{\footnotesize{Baseline}} & \textbf{\footnotesize{Prompt Template}} \\
\midrule

Direct Question &
You are a fact-checker. Respond only with `True' or `False'. \newline

Is the following claim true or false? Claim: \texttt{<claim>} \\

\midrule

Chain-of-Thought &
You are a careful reasoner. Think step by step before answering. \newline

Claim: \texttt{<claim>}. Instructions: Consider the claim and determine whether it is true or false. Think step by step before providing the final answer. Use critical thinking and logical reasoning. Your answer should conclude clearly with `True' or `False'. \\

\bottomrule
\end{tabular}
}
\caption{Prompt templates used in baselines. All models receive the same underlying instructions; only the formatting differs. Phi and Llama models use explicit inline tags, while Mistral and GPT-4o mini uses message-passing format (system/user roles).}
\label{tab:prompts}
\end{table}

\begin{table}[H]
    \centering
    \renewcommand{\arraystretch}{1.4}
    \scriptsize
    \begin{tabular}{p{0.22\linewidth} p{0.70\linewidth}}
    \toprule
    \textbf{\footnotesize{Model}} & \textbf{\footnotesize{Prompt Template}} \\
    \midrule
    
    Llama-3.1-8B-Inst. &
    You are an impartial analyst. Do not apologise or say “I'm sorry” in any response. \newline

    \textbf{\#\# Task} \newline
    Read the claim below and reason carefully about whether it is likely true or false. Then generate up to \texttt{\{num\_args\}} factually grounded and relevant arguments that either agree with or disagree with the claim. \newline
    \hspace*{0.05em}\texttt{-} If you judge the claim is more likely TRUE, output mostly "Agree" arguments. \newline
    \hspace*{0.05em}\texttt{-} If you judge the claim is more likely FALSE, output mostly "Disagree" arguments. \newline

    Only include arguments you are confident in. Avoid hedging, speculation, and vague language. \newline

    \textbf{\#\# Format} \newline
    Each argument must: \newline
    • Be one sentence under 30 words   \newline
    • Start exactly with "Agree:" or "Disagree:" \newline
    • Be self-contained: do not refer to “the claim” or restate it directly   \newline
    • Do not use bullets, numbering, or formatting \newline
    
    \textbf{\#\# Claim:} \newline
    \texttt{<claim>} \newline

    \textbf{\#\# Instructions:} \newline
    Think carefully. Then output up to \texttt{\{num\_args\}} arguments, one per line, in the format "Agree: ..." or "Disagree: ...". \\


    \bottomrule
    \end{tabular}
    \caption{Prompt template used to elicit arguments from Llama-3.1-8B-Inst.}
    \label{tab:argument-generation-prompts-llama}
    \end{table}

    \begin{table}[H]
        \centering
        \renewcommand{\arraystretch}{1.4}
        \scriptsize
        \begin{tabular}{p{0.22\linewidth} p{0.70\linewidth}}
        \toprule
        \textbf{\footnotesize{Model}} & \textbf{\footnotesize{Prompt Template}} \\
        \midrule
        
        Llama-3.1-8B-Inst. &
        You are a helpful, neutral reasoning assistant evaluating factual claims for research purposes. You are not providing medical advice. Do not apologise or add “I'm sorry…” in any response. \newline
    
        \textbf{\#\# Task} \newline
        Read the claim below and reason carefully about whether it is likely true or false. Then generate up to \texttt{\{num\_args\}} factually grounded and relevant arguments that either agree with or disagree with the claim. \newline
        \hspace*{0.05em}\texttt{-} If you judge the claim is more likely TRUE, output mostly "Agree" arguments. \newline
        \hspace*{0.05em}\texttt{-} If you judge the claim is more likely FALSE, output mostly "Disagree" arguments. \newline
    
        Only include arguments you are confident in. Avoid hedging, speculation, and vague language. \newline
    
        \textbf{\#\# Format} \newline
        Each argument must: \newline
        • Be one sentence under 30 words   \newline
        • Start exactly with "Agree:" or "Disagree:" \newline
        • Be self-contained: do not refer to “the claim” or restate it directly   \newline
        • Do not use bullets, numbering, or formatting \newline
        
        \textbf{\#\# Claim:} \newline
        \texttt{<claim>} \newline
    
        \textbf{\#\# Instructions:} \newline
        Think carefully. Then output up to \texttt{\{num\_args\}} arguments, one per line, in the format "Agree: ..." or "Disagree: ...". Do not apologise or add “I'm sorry…” in any response. \\

    
        \bottomrule
        \end{tabular}
        \caption{Prompt template used to elicit arguments from Llama-3.1-8B-Inst. This template is tailored for sensitive medical claims, used to reduce the likelihood of refusals by explicitly framing the task as non-advisory.}
        \label{tab:argument-generation-prompts-llama-medical}
        \end{table}

    \begin{table}[H]
        \centering
        \renewcommand{\arraystretch}{1.3}
        \scriptsize
        \begin{tabular}{p{0.22\linewidth} p{0.70\linewidth}}
        \toprule
        \textbf{\footnotesize{Model}} & \textbf{\footnotesize{Prompt Template}} \\
        \midrule
    
        Mistral-7B-Inst. & 
        You are an impartial analyst. Do not apologise or say “I'm sorry” in any response. \newline
    
        \textbf{\#\# Task} \newline
        Read the claim and consider both supporting and opposing perspectives. Then generate up to \texttt{\{num\_args\}} arguments total. \newline
    
        Each argument must be one sentence under 30 words. \newline
        Each must begin with either "Agree:" or "Disagree:" — nothing else. \newline
        Do NOT number the arguments. Do NOT use bullets. Do NOT group under headings like "Agree:" or "Disagree:" \newline
    
        \textbf{Claim:} \newline
        \texttt{<claim>} \newline
    
        Generate up to \texttt{\{num\_args\}} arguments TOTAL (not per stance), using only the following format: \newline
        Agree: \texttt{<argument>} \newline
        or \newline
        Disagree: \texttt{<argument>} \\
        
        \bottomrule
        \end{tabular}
        \caption{Prompt template used to elicit arguments from Mistral-7B-Inst.}
        \label{tab:argument-generation-prompts-mistral}
    \end{table}

    \begin{table}[H]
        \centering
        \renewcommand{\arraystretch}{1.3}
        \scriptsize
        \begin{tabular}{p{0.22\linewidth} p{0.70\linewidth}}
        \toprule
        \textbf{\footnotesize{Model}} & \textbf{\footnotesize{Prompt Template}} \\
        \midrule
    
        Phi-3-Mini-Inst.(4B) & 
        You are an impartial analyst. Do not apologise or add “I'm sorry…” in any response. \newline
    
        Read the claim below and reason carefully about whether it is likely true or false. Then generate up to \texttt{\{num\_args\}} factually grounded and relevant arguments that either agree with or disagree with the claim. If the statement is partially false or has to be interpreted in a very specific way to be considered true, consider it false. \newline
    
        \hspace*{0.05em}\texttt{-} If you judge the claim is more likely TRUE, output mostly "Agree" arguments. \newline
        \hspace*{0.05em}\texttt{-} If you judge the claim is more likely FALSE, output mostly "Disagree" arguments. \newline
    
        Only include arguments you are confident in. Avoid hedging, speculation, and vague language. \newline
    
        \textbf{\#\# Format} \newline
        Each argument must: \newline
        • Be one sentence under 30 words  \newline
        • Start exactly with "Agree:" or "Disagree:" \newline
        • Be self-contained: do not refer to “the claim” or restate it directly  \newline
        • Do not use bullets, numbering, or formatting \newline
        
        \textbf{\#\# Claim:} \newline
        \texttt{<claim>} \newline
    
        \textbf{\#\# Instructions:} \newline
        Think carefully. Then output up to \texttt{\{num\_args\}} arguments, ONE PER LINE, in the format "Agree: ..." or "Disagree: ...". \\

        
        \bottomrule
        \end{tabular}
        \caption{Prompt template used to elicit arguments from Phi-3-Mini-Inst.(4B)}
        \label{tab:argument-generation-prompts-phi}
    \end{table}
    
    \begin{table}[H]
        \centering
        \renewcommand{\arraystretch}{1.3}
        \scriptsize
        \begin{tabular}{p{0.22\linewidth} p{0.70\linewidth}}
        \toprule
        \textbf{\footnotesize{Model}} & \textbf{\footnotesize{Prompt Template}} \\
        \midrule
    
        Llama-3.1-8B-Inst. &
        You are an expert in argument mining. Given text with \texttt{<ACi>} argument tags, classify relations between components as either "support" or "attack". Only include pairs with a relation. \newline
        
        \textbf{\#\# Instruction} \newline
        For the text below, Return \texttt{JSON}: \newline
        \texttt{"list\_argument\_relation\_types": [[source, target, relation], ...]} \newline
        where each \texttt{relation} is either \texttt{"support"} or \texttt{"attack"}. \newline
    
        \texttt{<ac\_text>} \newline
    
        \textbf{\#\# Output} \newline
        \texttt{"list\_argument\_relation\_types": [[1, 2, "support"], [2, 3, "attack"]]} \\
        
     
        \bottomrule
        \end{tabular}
        \caption{Prompt template used for Argument Relation Identification and Classification (ARIC), using Llama-3.1-8B-Inst.}
        \label{tab:aric-relation-prompt}
    \end{table}

\begin{table}[H]
    \centering
    \renewcommand{\arraystretch}{1.4}
    \scriptsize
    \begin{tabular}{p{0.22\linewidth} p{0.70\linewidth}}
    \toprule
    \textbf{Scoring Target} & \textbf{Prompt Template} \\
    \midrule
    
    Claim &
    You are an expert reasoning assistant evaluating the truth of a claim. \newline
    \textbf{Task:} Carefully read the claim below. Think step by step using your factual knowledge and reasoning skills. Conclude with a truthfulness label that reflects how likely the claim is factually correct. \newline
    \textbf{Truthfulness Labels:} \newline
    \hspace*{0.05em}\texttt{- true} \newline
    \hspace*{0.05em}\texttt{- probably true} \newline
    \hspace*{0.05em}\texttt{- uncertain} \newline
    \hspace*{0.05em}\texttt{- probably false} \newline
    \hspace*{0.05em}\texttt{- false} \newline
    \textbf{Claim:} \texttt{"<claim>"} \\
    
    \midrule
    
    Argument &
    You are an expert reasoning assistant evaluating individual arguments in the context of a factual claim. \newline
    \textbf{Task:} Read the claim and the associated argument below. \newline
    \hspace*{0.05em}\texttt{-} Evaluate how directly the argument addresses the claim. \newline
    \hspace*{0.05em}\texttt{-} Consider whether the argument is factually grounded, or relies on assumptions or anecdotal evidence. \newline
    \hspace*{0.05em}\texttt{-} Assess ambiguity, generalisation, or irrelevance that may weaken the argument. \newline
    Then, select a certainty label from the list below that reflects how persuasive the argument is. Do not explain your answer. \newline
    \textbf{Certainty Labels:} \newline
    \hspace*{0.05em}\texttt{- certain} \newline
    \hspace*{0.05em}\texttt{- almost certain} \newline
    \hspace*{0.05em}\texttt{- quite certain} \newline
    \hspace*{0.05em}\texttt{- moderately certain} \newline
    \hspace*{0.05em}\texttt{- slightly certain} \newline
    \hspace*{0.05em}\texttt{- almost uncertain} \newline
    \hspace*{0.05em}\texttt{- uncertain} \newline
    \textbf{Claim:} \texttt{"<claim>"} \newline
    \textbf{Argument:} \texttt{"<argument>"} \newline
    \textbf{Output:} \texttt{Certainty: <your chosen label>} \\
    
    \bottomrule
    \end{tabular}
    \caption{Prompt templates used in estimated base score scoring, using GPT-4o mini.}
    \label{tab:estimated-score-prompts}
    \end{table}

    \begin{table}[H]
        \centering
        \renewcommand{\arraystretch}{1.3}
        \scriptsize
        \begin{tabular}{p{0.20\linewidth} p{0.32\linewidth} c}
        \toprule
        \textbf{Scoring Target} & \textbf{Label} & \textbf{Mapped Score} \\
        \midrule
        Claim & true & 0.95 \\
              & probably true & 0.75 \\
              & uncertain & 0.50 \\
              & probably false & 0.25 \\
              & false & 0.05 \\
        \midrule
        Argument & certain & 0.95 \\
                 & almost certain & 0.80 \\
                 & quite certain & 0.65 \\
                 & moderately certain & 0.50 \\
                 & slightly certain & 0.35 \\
                 & almost uncertain & 0.20 \\
                 & uncertain & 0.05 \\
        \bottomrule
        \end{tabular}
        \caption{Label-to-score mapping used to convert GPT-4o mini outputs into numerical base scores for claim and argument nodes.}
        \label{tab:label-score-mapping}
        \end{table}

\section{Additional Evaluation Results}
\label{app:additional_experiment}

Table \ref{tab:baseline-results} reports the performance for single LLM, ensemble of single LLMs, and GPT-4o-mini baselines using direct questioning and CoT prompting, respectively. In most cases, CoT achieves a higher accuracy.

\begin{table}[ht]
    \centering
    \begin{tabular}{cccc}
    \toprule
    \textbf{Dataset} & \textbf{Model} & \textbf{Direct} & \textbf{CoT} \\
    \midrule
    \multirow{5}{*}{\shortstack[c]{Truthful\\Claim}}
    & Llama-3.1-8B-Inst. & 0.650 & \textbf{0.710} \\
    & Mistral-7B-Inst. & \textbf{0.702} & 0.694 \\
    & Phi-3-Mini-Inst.(4B) & \textbf{0.656} & 0.636 \\
    & Ensemble of LLMs & 0.682 & \textbf{0.724} \\ 
    & GPT-4o-mini & 0.796 & \textbf{0.802} \\
    \midrule
    \multirow{5}{*}{\shortstack[c]{Strategy\\Claim}}
    & Llama-3.1-8B-Inst. & 0.580 & \textbf{0.652} \\
    & Mistral-7B-Inst. & 0.598 & \textbf{0.604} \\
    & Phi-3-Mini-Inst.(4B) & 0.594 & \textbf{0.612} \\
    & Ensemble of LLMs & 0.596 & \textbf{0.642} \\
    & GPT-4o-mini & 0.770 & \textbf{0.794} \\
    \midrule
    \multirow{5}{*}{\shortstack[c]{Med\\Claim}}
    & Llama-3.1-8B-Inst. & 0.554 & \textbf{0.564} \\
    & Mistral-7B-Inst. & 0.544 & \textbf{0.554} \\
    & Phi-3-Mini-Inst.(4B) & \textbf{0.602} & 0.576 \\
    & Ensemble of LLMs & 0.558 & \textbf{0.570} \\
    & GPT-4o-mini & 0.736 & \textbf{0.760} \\
    \bottomrule
    \end{tabular}
    \caption{Test accuracy of baseline prompting strategies for Llama 3.1 8B, Mistral, Phi-3-mini, ensemble of these three LLMs, and GPT-4o mini. The best performing method for each dataset-model combination is indicated in bold.}
    \label{tab:baseline-results}
    \end{table}

\begin{table*}[h]
    \centering
    \renewcommand{\arraystretch}{1.1}
    \begin{tabular}{wc{2.2cm}wc{2.2cm}wc{2cm}wc{1.3cm}wc{1.3cm}wc{1.3cm}wc{1.3cm}}
        \toprule
        \multirow{2}[2]{*}{\textbf{Dataset}} & \multirow{2}[2]{*}{\textbf{Framework}} & \multirow{2}[2]{*}{\textbf{Model}} &
        \multicolumn{2}{c}{\textbf{Estimated Args}} &
        \multicolumn{2}{c}{\textbf{Estimated All}} \\
        \cmidrule(lr){4-5} \cmidrule(lr){6-7}
        & & & \textbf{D=1} & \textbf{D=2} & \textbf{D=1} & \textbf{D=2} \\
        \midrule

        \multirow{3}[3]{*}{\shortstack{Truthful \\ Claim}} 
        & \multirow{2}[1]{*}{ArgLLM}     & Llama-3-8B-Inst. & 0.63 & 0.61 & 0.68 & 0.69 \\
        &                                & Mistral-7B-Inst.  & 0.65 & 0.67 & 0.76 & 0.75 \\
        \cmidrule(){2-7}
        & MArgE & Ensemble\textsuperscript{†}  & \textbf{0.69}  & \textbf{0.70} &  \textbf{0.84}  &  \textbf{0.84} \\

        \midrule

        \multirow{3}[3]{*}{\shortstack{Strategy \\ Claim}} 
        & \multirow{2}[1]{*}{ArgLLM}     & Llama-3-8B-Inst. & 0.54 & 0.53 & 0.61 & 0.58 \\
        &                                & Mistral-7B-Inst.    & 0.58 & 0.61 & 0.62 & 0.60 \\
        \cmidrule(){2-7}
        & MArgE & Ensemble   &   \textbf{0.69}   &  \textbf{0.69}    &   \textbf{0.81}   &   \textbf{0.82}   \\

        \midrule

        \multirow{3}[3]{*}{\shortstack{Med \\ Claim}} 
        & \multirow{2}[1]{*}{ArgLLM}     & Llama-3-8B-Inst. & 0.51 & 0.53 & 0.53 & 0.52 \\
        &                                & Mistral-7B-Inst.    & 0.50 & 0.51 & 0.53 & 0.52 \\
        \cmidrule(){2-7}
        & MArgE & Ensemble   &  \textbf{0.57}   &   \textbf{0.58}   &   \textbf{0.70}  &   \textbf{0.73}   \\

        \bottomrule
    \end{tabular}
    \caption{
        Accuracy comparison between ArgLLM and MArgE at depth=1 and depth=2 under both Estimated Args and Estimated All scoring schemes. \textsuperscript{†}The term “Ensemble” refers to combining argument trees generated by Llama-3.1-8B-Inst., Mistral-7B-Inst., and Phi-3-Mini-Inst.(4B).
    }
    \label{tab:argllm-vs-multiargllm}
\end{table*}

In Table \ref{tab:argllm-vs-multiargllm}, we compare MArgE against ArgLLM. The ArgLLM results are taken from the original paper \cite{Freedman_Dejl_Gorur_Yin_Rago_Toni_2025}, instantiated with different prompting strategies and argument quality scoring approaches from MArgE. Using a single model, ArgLLM only improves over the single LLM baseline (in Table \ref{tab:baseline-results}) on TruthfulClaim. As discussed in Section \ref{sec:main}, ArgLLM's accuracy is limited by its rigid argument generation process and the use of verbalised confidence scoring (prompting the argument-generating LLMs for how confident they are about each argument) as the base scores. MArgE significantly outperforms ArgLLMs in every configuration setup. Note that in Section \ref{ssec:accuracy_results}, MArgE also outperforms an ensemble of ArgLLMs instantiated using the same models and the same prompting strategy as in MArgE.

\section{Additional Ablation Studies}
\label{app:additional_ablation}

We explore two configurations of MArgE in addition to the approach described in Section \ref{sec:main}. 

First, in Step 2, after meshing the argument trees from each LLM to a single one, we perform argument mining (AM) on the merged tree to further discover attack and support relations. The rationale behind this exploration is that while argument generation captures local pros and cons effectively, it can miss cross-cutting rebuttals or reinforcing relations, especially when (1) relevant arguments reside in different subtrees or (2) the top-down structure leaves leaf nodes without incoming edges, preventing them from being challenged or supported. We prompt the Llama-3.1-B-instruct for finding such new relations, and report the ablation results in Table \ref{tab:ablation-am}. While showing comparable performance to the default MArgE on Truthful Claim dataset, AM does not improve performance on Strategy Claim dataset. 

We explore another argumentative semantics, Quadratic Energy Model (QEM)~\cite{potyka2018}, for Step 4 updating argument strength. Unlike DF-QuAD, QEM propagates argument strengths using an energy-based balancing of supports, attacks, and base scores. The comparison results are shown in Table \ref{tab:ablation-semantics}. We observe that DF-QuAD consistently outperforms QEM across most settings, aligning with prior findings from ArgLLM \cite{Freedman_Dejl_Gorur_Yin_Rago_Toni_2025}. 

\begin{table}[ht]
    \centering
    \begin{tabular}{ccccc}
    \toprule
    \textbf{Dataset} & \textbf{Config} & \textbf{Default} & \textbf{With AM} \\
    \midrule
    \multirow{4}{*}{\shortstack[c]{Truthful\\Claim}}
    & D=1, Est.Arg & \textbf{0.690} & \textbf{0.690} \\
    & D=1, Est.All  & 0.820 & \textbf{0.824} \\
    & D=2, Est.Arg & \textbf{0.690} & 0.670 \\
    & D=2, Est.All  & 0.820 & \textbf{0.832} \\
    \midrule
    \multirow{4}{*}{\shortstack[c]{Strategy\\Claim}}
    & D=1, Est.Arg & \textbf{0.660} & 0.632 \\
    & D=1, Est.All  & \textbf{0.796} & 0.790 \\
    & D=2, Est.Arg & \textbf{0.672} & 0.646 \\
    & D=2, Est.All  & \textbf{0.824} & 0.814 \\
    \bottomrule
    \end{tabular}
    \caption{Effect of augmenting merged trees with LLM-inferred support/attack edges using Argument Mining (AM)~\cite{cabessa2025}. Best performance for each dataset-configuration pair is bolded. Results shown are for the semantic merging strategy only.}
    \label{tab:ablation-am}
\end{table}

\begin{table}[ht]
    \centering
    \begin{tabular}{ccccc}
    \toprule
    \textbf{Dataset} & \textbf{Config} & \textbf{DF-QuAD} & \textbf{QEM} \\
    \midrule
    \multirow{4}{*}{\shortstack[c]{Truthful\\Claim}}
    & D=1, Est.Arg & 0.690 & \textbf{0.692} \\
    & D=1, Est.All  & \textbf{0.820} & 0.764 \\
    & D=2, Est.Arg & \textbf{0.690} & 0.676 \\
    & D=2, Est.All  & \textbf{0.820} & 0.764 \\
    \midrule
    \multirow{4}{*}{\shortstack[c]{Strategy\\Claim}}
    & D=1, Est.Arg & \textbf{0.660} & 0.656 \\
    & D=1, Est.All  & \textbf{0.796} & 0.768 \\
    & D=2, Est.Arg & \textbf{0.672} & \textbf{0.672} \\
    & D=2, Est.All  & \textbf{0.824} & 0.766 \\
    \bottomrule
    \end{tabular}
    \caption{Comparison of test accuracy using DF-QuAD versus the Quadratic Energy Model (QEM)~\cite{potyka2018} for gradual argument strength propagation. Best performance for each dataset-configuration pair is bolded. Results shown are for the semantic merging strategy only.}
    \label{tab:ablation-semantics}
\end{table}




\end{document}